\documentclass{article}

\usepackage{booktabs} 
\usepackage{amsmath,amssymb}
\usepackage{microtype}
\usepackage{multirow}
\usepackage{multicol}
\usepackage{graphicx}
\usepackage{subcaption}
\usepackage{amsfonts}
\usepackage[dvipsnames]{xcolor}
\usepackage{hyperref}
\usepackage{tabularx}
\usepackage[font=small,labelfont=bf]{caption}

\DeclareMathOperator*{\argmax}{arg\,max}
\DeclareMathOperator*{\argmin}{arg\,min}

\PassOptionsToPackage{numbers, sort&compress}{natbib}

\usepackage[preprint]{corl_2021} 

\title{S4RL: Surprisingly Simple Self-Supervision for Offline Reinforcement Learning in Robotics}

\author{
Samarth Sinha$^{1}$, 
Ajay Mandlekar$^{2}$,
Animesh Garg$^{1,3}$
\\
$^{1}$University of Toronto, Vector Institute,
$^{2}$Stanford University,
$^{3}$Nvidia
}

\begin{document}
\maketitle


\begin{abstract}
Offline reinforcement learning proposes to learn policies from large collected datasets
without interacting with the physical environment. 
These algorithms have made it possible to learn useful skills from data 
that can then be deployed
in the environment in real-world 
settings where interactions may be costly or dangerous, such as autonomous driving or factories.
However, current algorithms overfit to the dataset they are trained on and 
exhibit poor out-of-distribution generalization to the environment when deployed. 
In this paper, we study the effectiveness of performing data augmentations 
on the state space, and study 7 different augmentation schemes and how 
they behave with existing offline RL algorithms.
We then combine the best data performing augmentation scheme with a state-of-the-art 
Q-learning technique, and improve the function approximation of the Q-networks by smoothening out the learned state-action space.
We experimentally show that using this Surprisingly Simple Self-Supervision technique in RL (S4RL), we significantly improve over the current state-of-the-art algorithms on offline robot learning environments
such as MetaWorld \cite{metaworld} and RoboSuite \cite{robosuite,mandlekar2018roboturk},
and benchmark datasets such as D4RL \cite{d4rl}.
\end{abstract}

\keywords{Offline Reinforcement Learning, Data Augmentation, Self-Supervised Learning} 


\section{Introduction}
	
In reinforcement learning (RL), an agent is trained to interact with its environment and
learn useful skills that help with solving the given task. 
However, interacting with the environment may be costly, or 
unsafe to do from scratch in real-world scenarios such as self-driving or 
industrial robotics. 
Unlike directly interacting with the environment, it is much cheaper to collect 
and store data such that an agent can be trained on it and learn from the 
past experience.
In direct contrast to reinforcement learning, offline RL
(or batch RL) \cite{levine2020offline} allows a setup where a 
behavioral policy is used to collect and store the data such that a target policy
can then be used to train on the data, without any further interactions with the environments.
The behavioral policy can range from human demonstrations, random policy in the given
environment, or near-optimal policies for the task. 
However, learning from such demonstrations is a difficult task since the 
data does not cover the full state-action space, and naive behavioral policies will
not cover the state-action distribution for an optimal policy for a task.


Current offline model-free RL algorithms include learning a $Q$-function \cite{qlearning}
where a parameterized
neural network is trained to learn the state-action values from
data \cite{bcq, bear, d4pg, brac, agarwal2020optimistic,mandlekar2020iris}.
This family of algorithms suffer from overestimating the true state-action values
of data that is not in the same distribution as the offline dataset used.
Along with overestimation error, another source of error is with the function 
approximation of the neural networks that are typically used to parameterize
the $Q$-functions. 
Since the state-action distribution is static during training, the neural networks
may overfit the data, further resulting in poor generalization when deployed
to real-world settings.
To discourage overfitting in neural networks, recent pixel-based methods
have been proposed to utilize data augmentation to the image observations, 
resulting in state-of-the-art results for normal reinforcement learning 
\cite{curl, drq, rad}.
However, it is unclear to how perform such augmentations from proprioceptive 
information and the role of augmentation in offline RL.
In this paper we investigate the role of data augmentations from 
proprioceptive observations focused on offline RL.
Data augmentation also help perform better
function approximation for $Q$-networks since it allows to smoothen out the 
state space by ``visiting'' the local regions and ensuring the learned 
$Q$-values are similar throughout, since small perturbations in the 
given observation should not lead to drastically different $Q$-values.

Our main contribution is a Surprisingly Simple Self-Supervised offline RL (S4RL)
algorithm that combines studying data augmentations with a simple $Q$-learning 
method to significantly improve the performance of offline RL algorithm.
The proposed framework is general and can be added to a number of off-the-shelf offline/batch RL
algorithms for robot learning.
We present comparisons our model to other comparable self-supervised
learning strategies and state-of-the-art offline RL methods on
the D4RL benchmark dataset \cite{d4rl}, which consists of continuous control,
navigation and robotic control tasks collected from suboptimal and human agents.
We further experiment with two dexterous robot manipulation
environments: MetaWorld \cite{metaworld} and RoboSuite \cite{robosuite,mandlekar2018roboturk}.
Across all evaluations, we find that our proposed state-based data augmentation for $Q$-learning significantly 
outperforms base offline RL algorithm as well as various competitive data augmentation baselines. 


\begin{figure}[t!]
    \centering
    \includegraphics[width=0.75\textwidth]{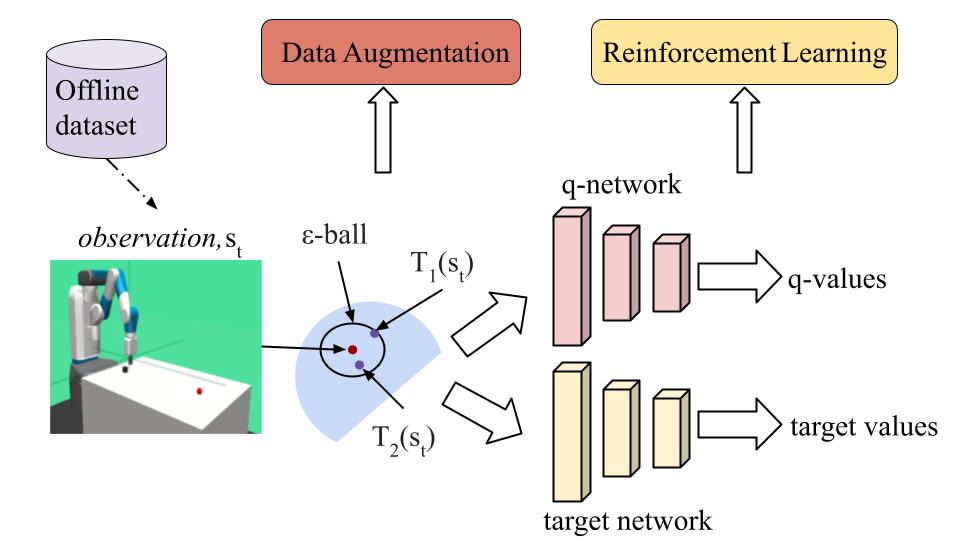}
    \caption{Overview of the proposed augmentation scheme. Here $T_1(s_t)$ and
    $T_2(s_t)$ represent different augmentation schemes. 
    We simply perform $k-$ data augmentations to the observations $s_t$, and
    use that the augmented states to calculate the q-values and target values
    for better function approximation of the Q-network by ensuring that local
    perturbations in the state space have similar $Q$-value estimates.}
    \label{fig:teaser}
\end{figure}

\section{Related Work}



\textbf{Offline reinforcement learning:}
In offline RL, a static dataset of demonstrations is collected, and is used
to train an agent for a given environment.
Popular actor-critic algorithms, such as Soft Actor Critic (SAC) \cite{sac},
tend to perform poorly on learning from offline datasets, 
because they are unable to generalize
to out-of-distribution (OOD) data because of an overestimation bias: where 
the critic overestimates the values of state-action pairs not previously encountered
from the collected data which leads to a brittle policy.


The convergence properties of value functions have been proven in simple settings
\cite{ormoneit2002kernel}; however learning a policy
using the learned value functions remains an active challenge in offline RL.
Simple algorithms like fitted $Q$-iterations have been able to learn state-action
value functions, but do not work well with sub-optimal data 
\cite{ernst2005tree, riedmiller2005neural}.
To help solve the generalization and extrapolation problem, different offline RL
algorithms have proposed to use constrained policy optimization between the 
target policy and the behaviour policy that was used to collect the data.
These constrains include using KL-divergence \cite{brac, jaques2019way}, MMD \cite{bear} or $f$-divergence \cite{lfiw}.
Some algorithms have also proposed to learn a state conditioned VAEs to 
minimize the \textit{extrapolation error} \cite{bcq, bear, ajay2020opal}.
Recent work by \citet{fdpo} investigated using worst-case analysis to train 
a policy, denoted as \textit{pessimism}.
Conservative $Q$-Learning (CQL) explicitly deals with the problem of
overestimation for OOD state-action distribution by \textit{pushing down}
the value of randomly sampled state-action distributions in the 
critic objective \cite{cql}.
Learning a dynamics model from offline data has been promising
\cite{mopo, morel}.
A recent review of offline RL outlines the state of the field \cite{levine2020offline}.
Recent work has also shown great promise using offline RL for real robotics
tasks by learning policies that can directly be deployed on real robots to 
perform complex tasks \cite{cog, cabi2019framework, jin2020offline}.
Similar to before, such work relies on collecting large scale robotic datasets
 \cite{roboturk,mandlekar2018roboturk,robonet},
and learning policies that can then be safely deployed after a few steps of
finetuning.

\textbf{Representation learning in RL:}
Recent work in state and pixel-based RL suggests the need to learn better
representations from data usig better network architectures \cite{d2rl} or
self-supervision \cite{curl}.
Self-supervised representation learning has recently been applied to learn RL 
agents from pixel-data.
By using data augmentations and contrastive learning, \citet{curl} showed 
significant improvements on learning from pixel data.
The need for contrastive learning was simplified by RL with Augmented Data (RAD)
\cite{rad} and Data-Regularized $Q$-learning (DrQ) \cite{drq} as they 
provide a simpler alternative that used data augmentations without a 
contrastive objective.
More recently, Self-Predictive Representations (SPR) optionally uses data
augmentations for a self-supervision \cite{spr}.
The closest work to S4RL include DrQ \cite{drq} and RAD \cite{rad} in that
we utilize their method for $Q$-learning over augmentations, but in contrast we focus 
specifically on offline RL for robotic tasks and propose augmentations from states; we also
benchmark different self-supervision techniques from states.

\section{Preliminaries}

In reinforcement learning, an agent interacts with an environment to learn an optimal
policy.
This is typically framed as a Markov Decision Process (MDP) which can be represented
by a tuple of the form ($\mathcal{S}$, $\mathcal{A}$, $\rho$, $r$, $\gamma$, $\rho_0$), 
where $\mathcal{S}$ is the state space, $\mathcal{A}$ is the action space,
$\rho(s_{t+1}|s_t,a_t)$ is the 
transition function given the current state and action pair, $r(s_t)$ is the 
reward model given the state, $\gamma$ is the discount factor where
$\gamma \in [0,1)$ and $\rho_0$ is the initial state distribution of the MDP.

A policy, represented by $\pi : \mathcal{S} \rightarrow \mathcal{A}$ is trained to maximize the expected cumulative
discounted reward in the MDP. 
This can simply be formalized as: $\mathbb{E} \Big[ \sum_{t=0}^{\infty} \gamma^t r(s_t, \pi(a_t|s_t)) \Big]$


Furthermore, a state-action value function, $Q(s_t,a_t)$, is the value of 
performing a given action $a_t$ given the state $s_t$.
The $Q$-function is trained by minimizing the Bellman Error over $Q$ in a step
called policy evaluation

\begin{equation}
   Q_{i+1} \leftarrow \argmin_Q \mathbb{E} \Big [r_t + \gamma Q_{i}(s_{t+1}, a_{t+1}) - Q_{i}(s_t,a_t) \Big]
   \label{bellmanerror}
\end{equation}

where $Q_i$ is the $i$-th step of policy evaluation.
The policy is then trained to maximize the state-action values of performing 
an action $a_t$ given $s_t$ in a step called policy improvement:

\begin{equation}
    \pi_{i+1} \leftarrow \argmax_\pi \mathbb{E} \Big [  Q(s_t,\pi_i(a_t|s_t))  \Big]
    \label{eqn:policy_eval}
\end{equation}
where $\pi_i$ is the $i$-th step of policy improvement.

Unlike traditional reinforcement learning, in offline RL, the goal is not to learn 
an optimal policy for the MDP, but rather to learn an optimal policy given the 
dataset.
A behaviour policy $\mu$ is used to collect a static dataset 
$\mathcal{D}$ which is then used to train a target policy $\pi$. 
The policy improvement can now be stated as

\begin{equation}
    \pi_{i+1} \leftarrow \argmax_\pi \mathbb{E}_{s_t\sim\mathbb{D}} \Big [  Q(s_t,\pi_i(a_t|s_t))  \Big]
    \label{eqn:policy_imp}
\end{equation}

where $\mathcal{D}$ is the static dataset.
Since the nature of the behaviour policy $\mu$ is unknown, and can be composed 
of one (or more) sub-optimal policies, the RL task becomes challenging.
Since offline RL algorithms tend to generalize poorly to OOD data, the nature
of $\mu$ and the optimality of $\mu$ is important.
As shown \citet{d4rl}, typical offline RL methods perform poorly when a mixture
of suboptimal policies or an untrained random policy is used to collect the dataset 
$\mathcal{D}$.
In deep reinforcement learning, both the policy and the value functions
are parameterized using neural networks and trained using gradient descent.

\begin{figure}[t!]
    \centering
    \includegraphics[width=0.6\textwidth]{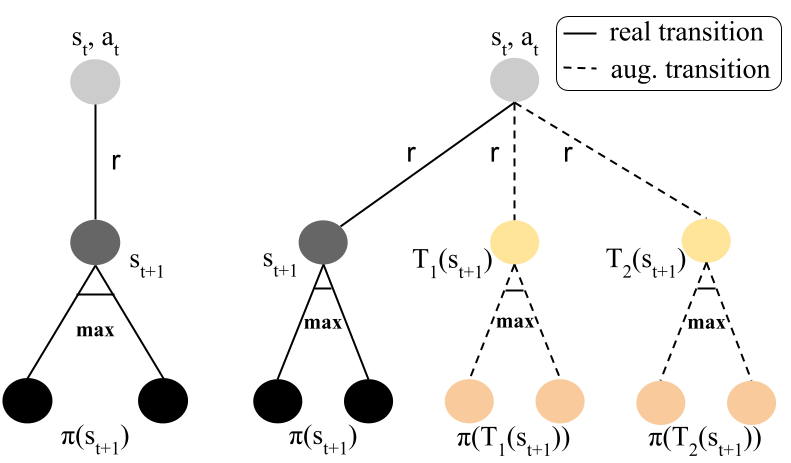}
    \caption{A diagram illustrating the difference between a Bellman 
    backup for offline RL versus offline RL with augmentations.
    Using data augmentations, we are able smoothen out the local state-space 
    around the observations in the dataset $s_t$.}
    \label{fig:bellman_backup}
\end{figure}

\section{Method}
In this section we will first consider a simple $Q$-learning technique based on
DrQ \cite{drq}, which helps combine data augmentations and learning 
better value functions during training by encouraging local points
around the state to have similar $Q$-values.
Then, we introduce 7 different data augmentations strategies 
that can be leveraged to allow the offline agent to perform local exploration
on the otherwise static dataset $\mathcal{D}$ to reduce overfitting and
improve function approximation by smoothening the state-space.

\subsection{Self-Supervision in Offline RL}
\label{sec:s4rl}

We denote a data augmentation transformation as: $\mathcal{T}(\tilde{s_t}|s_t)$
where $s_t \sim \mathcal{D}$, $\tilde{s_t}$ is the augmented state, and
$\mathcal{T}$ is a stochastic transformation.
After producing one or multiple augmentations of a given state 
$\tilde{s_t} \sim \mathcal{T}(s_t)$, we need a method to incorporate the augmentation to encourage smoothness and consistency.
For example, \cite{curl} uses contrastive learning to ``pull'' the representations
of two augmentations of the same state towards each other, 
while ``pushing'' two different state representations apart from each other.
We follow a similar scheme to DrQ \cite{drq}, where we perform multiple
augmentations of a given state, and we simple average the state-action values and
target values over the different augmetations of the state.

\begin{equation}
      \min_Q \mathop{\mathbb{E}}_{s_t,a_t \sim \mathcal{D}} \Big [r_t + \gamma \frac{1}{i} \sum_i Q(\mathcal{T}_i(\tilde{s}_{t+1}|s_{t+1}), a_{t+1}) - 
      \frac{1}{i} \sum_i Q(\mathcal{T}_i(\tilde{s_t}|s_t),a_t) \Big]
      \label{eqn:new_obj}
\end{equation}

The main difference between our proposed objective and the normal
$Q$-learning objective is
the mean over $i$ different augmentations in the second term of the equation. 
We augment the Bellman error in Equation \ref{bellmanerror} to simply be the 
mean error over the $i$ different augmentations of the same state.
Intuitively, this will help improve the consistency of the $Q$-value
within some perturbation field of the current state $s_t$, since the 
Bellman backups are taken over $i$-different views of the same state $s_t$.

By simply averaging the state-action values and target values, we assume
that the reward function is locally smooth to small perturbations in the
state.
Utilizing multiple augmentations further allows the networks to learn different 
variations of the data; we explore the role of increasing the number of different
augmentations empirically in Section \ref{sec:results}.
We use the trained $Q$-networks to perform policy improvement using the objective
in Equation \ref{eqn:policy_imp}, without augmentations.
The benefits of the augmentations are distilled to the policy since the value
functions are used to train the policy.
By combining such local perturbations to the states with 
self-supervision, we are able to learn a more robust policy that
can generalize better to unseen data when deployed on robots.
An overview of the proposed method is available in Figure \ref{fig:teaser},
where we disentangle the data augmentation and reinforcement learning
steps to learn $q$-values and target values over $i$ different augmentations
of the original state.
Without loss of generality, to show local perturbations, we draw an 
$\epsilon$-ball over the current state $s_t$, however different versions of 
augmentations are possible, as will be discussed.

\subsection{Data Augmentations}
\label{sec:data_aug}
In computer vision research, augmentation are commonly used as a way to obtain
the same datapoint from multiple viewpoints.
Transformations such as rotation, translations, color jitters, etc. are commonly 
used to train neural networks. 
Such transformations preserve the semantics of the image after the transformation
since for example: an image of a cat rotated $15^\circ$ remains an image of a cat.
However, when working from only proprioceptive information of an agent (for
example: the joint angles and velocity information of an industrial robot), such 
transformations are semantically meaningless.

In offline RL, since an agent is unable to interact with the environment to 
collect new data samples, data augmentations serve as a simple technique that
can allow the agent to do local exploration from the trajectories in the 
dataset.
Data augmentations from states, assumes that the output of a small transformation 
to an input state results in a physically realizable state in the environment.
If the augmented state is in the domain of the state space, 
$\tilde{s_t} \in \mathcal{S}$, then we consider it a valid transformation.
By performing valid augmentation stratgies, we are able to artificially 
increase the amount of data available during training, thereby reducing
the chances of the overfitting to the data.

Even if a transformation is physically realizable in the environment, a 
data augmentation strategy that is too aggressive may end up hurting the 
RL agent since the reward for the original state may not coincide with the
reward obtained from the augmented state. 
That is $r(s_t) \neq r(\tilde{s_t})$.
Therefore, one key assumption made when performing data augmentations
on the states is that the reward model $r(s_t)$ is \textit{smooth}, that is
$r(s_t) \approx r(\tilde{s_t})$.
Therefore the choice of $\mathcal{T}(\tilde{s_t}|s_t)$ needs to be 
a local transformation to perturb the state without changing the semantics.
The choice of $\mathcal{T}(\tilde{s_t}|s_t)$ is more important in offline
RL than in traditional RL, since in traditional RL the agent may be able to 
self-correct for poor choices of transformations by 
being able to visit those states and correct the value functions using 
Bellman backups.
A comparison is presented in Figure \ref{fig:bellman_backup},
where we see that compared to normal offline RL, adding augmentation allows
for multiple paths in the policy evaluation step using Bellman backups.
By using data augmentation, we seek to exploit this smoothness property 
of the reward model in the MDP, which enables us to artificially visit 
states that may be in the state space, but not in the dataset:
$s_t,\tilde{s_t} \in \mathcal{S}$ but $\tilde{s_t} \notin \mathcal{D}$.

Following the constraints, we consider the following transformations
$\mathcal{T}(\tilde{s_t}|s_t)$:

\begin{enumerate}
    \item \textbf{zero-mean Gaussian noise} to the state: 
    $\tilde{s_t} \leftarrow s_t + \epsilon$, where $\epsilon \in \mathcal{N}(0, \sigma I)$
    and $\sigma$ is an important hyperparameter, which we will later discuss.
    \item \textbf{zero-mean Uniform noise} to the state: 
    $\tilde{s_t} \leftarrow s_t + \epsilon$, where $\epsilon \in \mathcal{U}(-\alpha, \alpha)$
    and $\alpha$ is an important hyperparameter.
    \item \textbf{random amplitude scaling} as in RAD \cite{rad}: $\tilde{s_t} \leftarrow s_t * \epsilon$, 
    where $\epsilon \in \mathcal{U}(\alpha, \beta)$, which preserves the 
    signs of the states.
    \item \textbf{dimension dropout} which simply zeros one random dimension in the state:
    $\tilde{s_t} \leftarrow s_t \cdot \mathbf{1}$ where $\mathbf{1}$ is a vector of 1s
    with one 0 randomly sampled from a Bernoulli distribution.
    This transformation preserves all-but-one intrinsic values of the current state.
    \item \textbf{state-switch} where we flip the value of 2 randomly selected dimensions 
    in the state. 
    This naive transformation will likely break the ``physical realizability'' assumption,
    since it is possible that the two randomly selected samples are semantically dissimilar
    properties of a robot (such as joint angle and joint velocity).
    To overcome this we hardcode the pairs of dimensions that can be switched
    for each environment (details in Appendix \ref{state-switch}).
    \item \textbf{state mix-up} where we use mixup \cite{mixup} between the current and
    the next state: $\tilde{s_t} \leftarrow \lambda s_t + (1-\lambda)s_{t+1}$, where
    $\lambda \sim \texttt{Beta}(\alpha, \alpha)$ with $\alpha = 0.4$ as in \cite{mixup}.
    The next state $s_{t+1}$ remains unchanged.
    \item \textbf{adversarial state training} which we take the gradient with
    respect to the
    value function to obtain a new state. This can be written as 
    $\tilde{s_t} \leftarrow s_t + \epsilon \nabla_{s_t} \mathbb{J}_Q(\pi(s_t))$, where $\mathbb{J}_Q$ is simply the policy evaluation update as in Equation \ref{eqn:policy_eval}, and $\epsilon$ constraints the size of the gradient. 
    The intuition behind this is to choose the direction within an 
    $\epsilon$-ball of $s_t$ where the $Q$-value deviates the most.
    This augmentation is intuitively the most useful in state spaces with high 
    dimensionality where random noise may be less effective \cite{mandlekar2017adversarially}.
\end{enumerate}


\begin{figure}
    \centering
    \includegraphics[width=0.9\columnwidth]{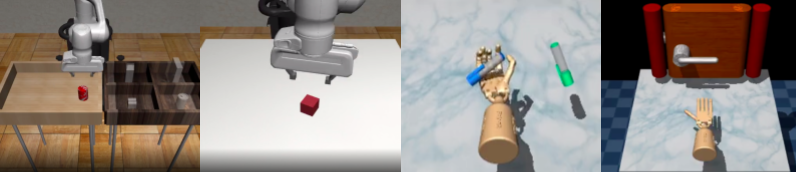}
    \caption{Illustration of the robotic environments used in the experiments. 
    The environments range from robotic arms for difficult manipulation tasks such as
    opening a door or placing a can into a bin or dexterous manipulation of a pen to a desired configuration using a robot hand.}
    \label{fig:env}
\end{figure}
\section{Experiments}
\label{sec:exp_setup}

In this section, we will first describe the dataset (D4RL \cite{d4rl}), and some 
hyperparameters and information about the experimental setup.
Then we will use S4RL to investigate the effect of 
different data augmentation strategies on the OpenAI Gym subset of D4RL tasks 
to investigate which augmentations are useful for learning.
Using the best data augmentation techniques found, we compare the algorithm to other 
self-supervision techniques that have been proposed for pixel-based RL, namely 
Contrastive Unsupervised Reinforcement Learning (CURL) 
\cite{curl}, and SAC+AutoEncoder (SAC+AE) where we feed the augmentations to the 
AutoEncoder to learn generalizable representations \cite{sac_ae}. 
We add S4RL framework to two 
state-of-the-art offline RL algorithms: Conservative $Q$-learning (CQL)
\cite{cql} and Behaviour Regularized Actor Critic (BRAC) \cite{brac} 
to show the effectiveness over both baseline agents to show the generality of
S4RL.
We add the CQL regularization to each self-supervised baseline considered,
since the self-supervised baselines were considered for online RL and CQL
adds a regularization term to help with overestimation.
We first test each baseline on the full suite of D4RL tasks \cite{d4rl}, 
which includes environments such as a Maze environment where an ant 
agent must learn locomotion and navigation skills to reach an environment goal 
which requires hierarchical control,
Adroit domain which requires learning dextrous object manipulation 
for fine-grained control of a robotic hand \cite{adroit}, 
Kitchen environment which simulates a Franka Panda robot to perform kitchen tasks.
We additionally experiment with more difficult robotic environments: 
MetaWorld \cite{metaworld} which uses a simulated Sawyer robot 
and RoboSuite \cite{robosuite} which uses a 
Franka Panda robotic arm for manipulation.
We further benchmark each of the baselines on \textit{learning with limited data}
in Appendix \ref{sec:limited_data},
to investigate the effect of self-supervision when the agent has significantly less
data to train from.
Finally, we experiment with different ways to regularize $Q$-network to prevent
overfitting, such as Dropout and Weight Decay in Appendix \ref{sec:regularization}
and include ablation experiments investigating the number of augmentations
and the role of hyperparameters in Appendix \ref{sec:hyperparam}. 



\paragraph{Hyperparameters:}
For all experiments, we do not perform any hyperparamter tuning to the base CQL
agent; all agents are trained using the original hyperparameters.
Similar to DrQ, we use 2 augmentations for obtaining the state-action values and 
the target values. 
We use a $\sigma$ and $\alpha$ value of $3 \times 10^{-4}$  for the zero-mean Gaussian and Uniform noise
augmentation variant, respectively.
We use a value of $1 \times 10^{-4}$ for $\epsilon$ for adversarial training; an
ablation on $\sigma$ and $\epsilon$
is available in Appendix \ref{sec:hyperparam}.


\begin{table}[t!]
\begin{small}
    \centering
    \resizebox{\columnwidth}{!}{%
        \begin{tabular}{l | c c c c c c c c}
        \toprule
         & \multicolumn{8}{c}{\textbf{CQL}} \\
         & & \textbf{+S4RL} & \textbf{+S4RL} & \textbf{+S4RL} & \textbf{+S4RL} & \textbf{+S4RL} & \textbf{+S4RL} & \textbf{+S4RL} \\
        \textbf{Task Name} & \textbf{Normal} & ($\mathcal{N}$) & ($\mathcal{U}$) & (Amp-Scale) & (Dim-Drop) & (State-Switch) & (MixUp) & (Adv) \\
        \midrule
            cheetah-random & 35.4 & \textbf{\textcolor{Sepia}{52.3}} & \textbf{\textcolor{OliveGreen}{50.3}} & 46.4 & 45.5 & 40.3 & 45.2 & \textbf{\textcolor{Blue}{53.9}} \\
            cheetah-medium & 44.4 & \textbf{\textcolor{Blue}{48.8}} & \textbf{\textcolor{OliveGreen}{47.0}} & 42.5 & 45.6 & 41.2 & 46.2 & \textbf{\textcolor{Sepia}{48.6}} \\
            cheetah-medium-replay & 42.0 & \textbf{\textcolor{Sepia}{51.4}} & \textbf{\textcolor{OliveGreen}{50.8}} & 43.1 & 49.6 & 41.0 & 46.2 & \textbf{\textcolor{Blue}{51.7}} \\
            cheetah-medium-expert & 62.4 & \textbf{\textcolor{Blue}{79.0}} & \textbf{\textcolor{Sepia}{78.5}} & 71.2 & 66.1 & 68.3 & 73.1 & \textbf{\textcolor{OliveGreen}{78.1}} \\
            \midrule
            hopper-random & \textbf{\textcolor{Sepia}{10.8}} & \textbf{\textcolor{Sepia}{10.8}} & 10.7 & \textbf{\textcolor{Sepia}{10.8}} & 9.3 & 9.5 & \textbf{\textcolor{Blue}{11.0}} & 10.7 \\
            hopper-medium & 58.0 & 78.9 & \textbf{\textcolor{Sepia}{80.6}} & 60.3 & 45.3 & 54.6 & \textbf{\textcolor{OliveGreen}{79.2}} & \textbf{\textcolor{Blue}{81.3}}\\
            hopper-medium-replay & 29.5 & \textbf{\textcolor{OliveGreen}{35.4}} & 35.2 & 28.6 & 20.4 & 20.9 & \textbf{\textcolor{Sepia}{35.6}} & \textbf{\textcolor{Blue}{36.8}} \\
            hopper-medium-expert & 111.0 & \textbf{\textcolor{Sepia}{115.2}} & 112.3 & 102.6 & 98.3 & 108.6 & \textbf{\textcolor{OliveGreen}{113.5}} & \textbf{\textcolor{Blue}{117.9}} \\
            \midrule
            walker-random & 7.0 & \textbf{\textcolor{Sepia}{24.9}} & \textbf{\textcolor{OliveGreen}{20.5}} & 6.9 & 3.2 & 9.9 & 10.2 & \textbf{\textcolor{Blue}{25.1}} \\
            walker-medium & 79.2 & \textbf{\textcolor{Sepia}{93.6}} & \textbf{\textcolor{Blue}{93.9}} & 81.6 & 65.3 & 65.3 & 88.6 & \textbf{\textcolor{OliveGreen}{93.1}} \\
            walker-medium-replay & 21.1 & \textbf{\textcolor{OliveGreen}{30.3}} & \textbf{\textcolor{Sepia}{31.9}} & 25.1 & 8.5 & 9.3 & 27.6 & \textbf{\textcolor{Blue}{35.0}} \\
            walker-medium-expert & 98.7 & \textbf{\textcolor{Blue}{112.2}} & 104.1 & 100.2 & 86.8 & 103.8 & \textbf{\textcolor{OliveGreen}{104.3}} & \textbf{\textcolor{Sepia}{107.1}} \\
            \midrule 
            \midrule
            \textbf{average-score} & 49.96 & \textbf{\textcolor{Sepia}{60.57}} & \textbf{\textcolor{OliveGreen}{59.65}} & 51.60 & 45.31 & 47.73 & 56.73 & \textbf{\textcolor{Blue}{61.6}} \\
            \textbf{average-ranking} & 5.83 & \textbf{\textcolor{Sepia}{1.92}} &\textbf{\textcolor{OliveGreen}{3.00}} & 5.00 & 6.42 & 6.25 & 3.17 & \textbf{\textcolor{Blue}{1.83}} \\
        \bottomrule
    \end{tabular}%
    }
    \caption{Investigating the importance of \textbf{different augmentation schemes in offline RL}. 
    We train a baseline CQL agent, and self-supervised CQL+S4RL agents using different
    augmentation schemes, and report the \textbf{mean normalized scores over 5 random seeds}.
    We report the baseline CQL results directly from \citet{d4rl}.
    To make the table easier to read we color the best algorithm on a task in \textcolor{Blue}{\textbf{blue}},
    the second best in \textcolor{Sepia}{\textbf{brown}}, and the third best in \textcolor{OliveGreen}{\textbf{green}}.
    We also report the average-score and average-ranking found using that augmentation scheme.
    }
    \label{tab:aug_table}
    \end{small}
\end{table}

\begin{table*}[t!]
    \centering
    \resizebox{\columnwidth}{!}{%
    \begin{tabular}{l|l | c  c c | c c c || c c c c}
        \toprule
        &  &  \multicolumn{6}{c||}{\textbf{CQL}} & \multicolumn{4}{c}{\textbf{BRAC-v}} \\
         & &  & \textbf{+CURL}  & \textbf{+VAE}  & \textbf{+S4RL} & \textbf{+S4RL} & \textbf{+S4RL} &  & \textbf{+S4RL} & \textbf{+S4RL} & \textbf{+S4RL} \\
         
        \textbf{Domain} & \textbf{Task Name} & \textbf{Normal} &  ($\mathcal{N}$) & ($\mathcal{N}$) & ($\mathcal{N}$) & (Mix-Up) & (Adv) & \textbf{Normal} & ($\mathcal{N}$) & (Mix-Up) & (Adv)\\
        \midrule
        \midrule
        \multirow{6}{*}{AntMaze} 
        & antmaze-umaze & 74.0 & 70.3 & 86.1 & \textbf{91.3} & 86.7 & \textbf{94.1} & 70.0 & 76.5 & 66.4 & \textbf{80.3} \\
        & antmaze-umaze-diverse & 84.0 & 81.4 & 85.1 & \textbf{87.8} & \textbf{86.9} & \textbf{88.0} & 70.0 & \textbf{81.3} & 65.4 & \textbf{80.9} \\
        & antmaze-medium-play & \textbf{61.2} & \textbf{60.9} & \textbf{62.1} & \textbf{61.9} & \textbf{61.3} & \textbf{61.6} & 0.0 & 0.0 & 0.0 & 0.0  \\
        & antmaze-medium-diverse & 53.7 & 45.4 & 59.9 & 78.1 & 62.3 & \textbf{82.3} & 0.0  & 0.0 & 0.0 & 0.0 \\
        & antmaze-large-play & 15.8 & 12.3 & 15.2 & \textbf{24.4} & 16.7 & \textbf{25.1} & 0.0 & 0.0 & 0.0 & 0.0 \\
        & antmaze-large-diverse & 14.9 & 8.3 & 17.3 & \textbf{27.0} & 15.9 & \textbf{26.2} & 0.0 & 0.0 & 0.0 & 0.0 \\
        \midrule
        \midrule
        \multirow{12}{*}{Gym} 
        & cheetah-random & 35.4 & 43.1 & 35.7 & \textbf{52.3} & 45.2 & \textbf{53.9} & 31.2 & \textbf{35.6} & 34.3 & \textbf{36.1} \\
        & cheetah-medium & 44.4 & 44.8 & 45.6 &  \textbf{48.8} & 46.2 & 48.6 & 46.3 & \textbf{49.1} & 46.1 & 46.0 \\
        & cheetah-medium-replay & 42.0 & 36.5 & 41.9 & 51.4 & 46.2 & \textbf{51.7} & \textbf{47.7} & \textbf{47.9} & \textbf{47.0} & \textbf{47.9} \\
        & cheetah-medium-expert & 62.4 & 65.6 & 70.7 & \textbf{79.0} &  73.1 & 78.1 & 41.9 & \textbf{52.1} & 46.7 &  \textbf{53.4} \\
        & hopper-random & \textbf{10.8} & \textbf{10.8} & \textbf{10.8} & \textbf{10.8} & \textbf{11.0} & 10.7 &  \textbf{12.2} & \textbf{12.9} & 10.9 & \textbf{12.1} \\
        & hopper-medium & 58.0 & 61.9 & 66.4 & \textbf{78.9} & \textbf{79.2} & \textbf{81.3} & 31.1 & 45.1 & 41.9  & \textbf{48.0} \\
        & hopper-medium-replay & 29.5 & 30.1 & \textbf{34.8} & \textbf{35.4} & \textbf{35.6} & \textbf{36.8} & \textbf{0.6} & \textbf{1.1} & \textbf{0.9} & \textbf{0.4} \\
        & hopper-medium-expert & \textbf{111.0} & \textbf{109.7} & \textbf{114.3} & \textbf{113.5} & \textbf{112.3} & 117.9 & 0.8 & \textbf{59.0} & 49.5 & 55.3 \\
        & walker-random & 7.0 & 6.6 & 6.5 & \textbf{24.9} & 10.2 & \textbf{25.1} & 1.9 & 9.5 & 2.4 & \textbf{14.1} \\
        & walker-medium & 79.2 & 81.3 & 80.8 & \textbf{93.6} & 88.6 & \textbf{93.1} & 81.1 & \textbf{85.4} & 80.9 & \textbf{86.7} \\
        & walker-medium-replay & 21.1 & 24.5 & 26.4 & 30.3 & 27.6 & \textbf{35.0} & \textbf{0.9} & \textbf{1.1} & \textbf{0.9} & \textbf{1.1} \\
        & walker-medium-expert & 98.7 & 103.0 & 99.6 & \textbf{112.2} & 104.3 & 107.1 & 81.6 & 90.4 & 89.7 &  \textbf{94.5} \\
        \midrule
        \midrule
        \multirow{8}{*}{Adroit} 
        & pen-human & 37.5 & 33.4 & 39.7 &  \textbf{44.4} & 41.6 & \textbf{51.2} & 0.6 & \textbf{11.3} & 0.8 & \textbf{14.1} \\
        & pen-cloned & 39.2 & 40.1 & 41.3 & \textbf{57.1} & 44.5 & \textbf{58.2} & -2.5 & \textbf{10.6} & 0.5 & \textbf{9.9} \\
        & hammer-human & 4.4 & \textbf{6.1} & \textbf{6.0} & \textbf{5.9} & \textbf{6.3} & \textbf{6.3} & 0.2 & \textbf{3.2} & \textbf{3.3} & \textbf{5.4} \\
        & hammer-cloned & 2.1 & 1.9 & 2.1 & \textbf{2.7} & \textbf{2.4} & \textbf{2.9} & 0.3 & 0.1 & 0.3 & \textbf{1.2} \\
        & door-human  & 9.9 & 10.3 & 14.3 & 27.0 & 16.7 & \textbf{35.3} & -0.3 & \textbf{2.8} & 0.2 & \textbf{4.2} \\
        & door-cloned  & 0.4 & 0.4 & 0.7 & \textbf{2.1} & 0.3 & 0.8 & -0.1 & \textbf{0.3} & -0.1 & \textbf{0.5} \\
        & relocate-human & \textbf{0.2} & \textbf{0.2} & \textbf{0.2} & \textbf{0.2} & \textbf{0.2} & \textbf{0.2} & -0.3 & \textbf{0.1} & \textbf{0.1} & \textbf{0.2} \\
        & relocate-cloned  & \textbf{-0.1} & \textbf{-0.1} & -0.3 & \textbf{-0.1} & \textbf{-0.1} & \textbf{-0.1} & -0.3 & \textbf{-0.1} & \textbf{-0.1} & \textbf{-0.1} \\
        \midrule
        \midrule
        \multirow{2}{*}{Franka} 
        & kitchen-complete & 43.8 & 54.4 & 42.1 & 77.1 & 60.4 & \textbf{88.1} & 0.0 & 0.0 & 0.0 & \textbf{12.3} \\
        & kitchen-partial & 49.8 & 59.6 & 49.9 & 74.8 & 59.0 & \textbf{83.6} & 0.0 & 0.0 &  0.0& \textbf{13.9} \\
        \bottomrule
    \end{tabular}%
    }
    \caption{\textbf{Full-set of experiments on the D4RL suite of tasks}.
    We perform experiments over 4 different domains and many different
    environments and data distributions for each environment.
    We compare two S4RL variants ($\mathcal{N}$ and Mix-Up) with two different 
    popular unsupervised representation learning methods and the base
    CQL offline RL agent.
    We train the CURL and VAE variants with the Gaussian additive
    noise variant that was empirically the most effective.
    We report the mean normalized performance over 5 random seeds.}
    \label{tab:main_table}
\end{table*}

\subsection{Results}
\label{sec:results}

\paragraph{Role of Data Augmentations}
We first investigate different data augmentation schemes and its performance
on the OpenAI Gym subset of the D4RL tasks.
We compare a base CQL agent with and without 
different forms of augmentation strategies, 
and tabulate the mean normalized performance, average normalized performance
over all the tasks and relative rankings in Table \ref{tab:aug_table}.
We see that using zero-mean Gaussian noise ($\mathcal{N}$), zero-mean
Uniform noise ($\mathcal{U}$), state mix-up and adversarial state training
consistently outperform the
the baseline CQL agent, as well as different data augmentation variants.
The average-ranking of S4RL-Adv and S4RL-$\mathcal{N}$ are 1.83 and
1.92 respectively, suggesting their effectiveness over a wide
range of task and data distributions.
We also see that CQL+S4RL agent is able to learn useful policies given
data collected from a random policy as evidenced by the performance
in ``walker-random'' where the base CQL agent is unable to learn any 
meaningful policy.

We also see that the S4RL agent is able to significantly outperform the 
baseline CQL agent on complex data distributions such as ``-medium-replay''
where the data collected is from all the data collected while training a 
policy in the environment.
Therefore the data split consists of data that ranges from a completely 
untrained policy (random policy), to a ``medium'' trained policy.

It is also important to look at the different augmentation techniques 
that do not help policy learning.
Techniques that hurt the performance and perform worse than the baseline
CQL agent include Dimension-Dropout and State-Switch.
Both techniques are inspired by popular computer vision data augmentation
algorithms, mainly MixMatch \cite{mixmatch} and CutMix \cite{cutmix}.
Since both techniques perform element-wise operations, it is likely that 
they omit important information about the state of the robot such as 
joint velocity.
Without such information, it is possible that the value function
is unable to reason about the environment which results in poor value estimates.

\paragraph{Comparison to other self-supervision techniques}


Following insights from Table \ref{tab:aug_table}, we choose three data
augmentation approaches, namely S4RL-$\mathcal{N}$, S4RL-MixUp and
S4RL-Adv, to compare to other self-supervision techniques.
We present the results over all the different tasks in D4RL in Table 
\ref{tab:main_table} using base CQL \cite{cql} and BRAC agents \cite{brac}.
Despite its simplicity, we continuously observe that +S4RL-$\mathcal{N}$ 
and +S4RL-adv agents
are able to significantly outperform the baselines on almost all tasks, and is 
comparable to the best in the others for both offline RL algorithms.
We specifically note that +S4RL-Adv significantly outperforms all other baselines
when the state-dimensionality is high, such as the dexterous manipulation 
environment of Adriot and on Franka robot.
In challenging environments that require hierarchical control such as the
AntMaze environments, we see that the S4RL agent continues to be the 
best performing agent.
Interestingly, on certain environments namely ``pen-cloned'', ``pen-human'', 
``hammer-human'' and ``door-human''
the BRAC+S4RL agent is able to learn some useful skills, whereas the base
CQL agent performs poorly. 
This further highlights the usefulness and generality of performing 
state-based augmentations for offline reinforcement learning.

\begin{figure}[t!]
    \centering
    \begin{subfigure}{0.47\textwidth}
    \includegraphics[width=\textwidth]{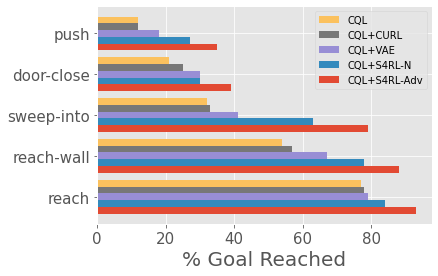}
    \caption{MetaWorld Environments}
    \end{subfigure}
    ~
    \begin{subfigure}{0.5\textwidth}
    \includegraphics[width=\textwidth]{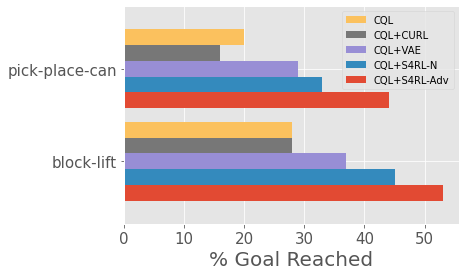}
    \caption{RoboSuite Environments}
    \end{subfigure}
    \caption{Results on Dexterous Manipulation Environments from MetaWorld \cite{metaworld} and RoboSuite~\cite{robosuite,mandlekar2018roboturk}. 
    We report the fraction (as \%) of goals reached in each of the environments during evaluation. Notably with a fixed amount of data, S4RL augmentation improves policy success rate over base CQL by approx. 20\% across different tasks. }
    \label{fig:robotic_exps}
\end{figure}

\paragraph{Dexterous Manipulation Environments}

Finally, we perform more experiments in the robotic domain using the MetaWorld 
\cite{metaworld} and RoboSuite environments on difficult manipulation 
tasks.
For MetaWorld, we train a SAC agent \cite{sac}
for 1M steps and collect 1000 trajectories of 200 episode length.
For RoboSuite, we collect data at 3 different instances during training to 
make the data distribution more complex.
More details on the data collection is available in Appendix 
\ref{sec:robosuite_data}.
The results are presented in Figure \ref{fig:robotic_exps}.
Similar to before we see significant improvements over the baseline in each of the 
7 environments over both domains.
The proposed S4RL-Adv agent is the only agent that is able to 
outperform the trained behaviour policy on 3 of the 5 MetaWorld tasks, 
while continuing to significantly outperform the baselines by as much as 3 times,
as in the case of ``push-v1'' and ``pick-and-place-can''.
Full numerical results can be found in Appendix \ref{sec:robotic_table}

\section{Conclusion}
In this paper, we present S4RL: a Surprisingly Simple Self-Supervised offline RL 
method that uses data augmentations to improve the function
approximation for $Q$-learning algorithms in offline RL. 
S4RL offers simplicity and ease of implementation, and can be added to any offline agent
that requries $Q$-learning.
We first compare and benchmark the effectiveness of 7 different data augmentation 
strategies from states in offline RL and 
then use the insights to compare against different self-supervised representation learning 
algorithms that have been proposed for pixel-based online RL.
We observe significant performance gains over the benchmark D4RL dataset \cite{d4rl}
and on more dexterous robot manipulation environments. We observe significant performance gains which make offline policy learning increasingly more competitive and help robot learning from only past data. 
Interesting future extension of this work can seek to use the suggested 
augmentation schemes to build better self-supervised learning algorithms.


\clearpage


{\small
\bibliography{example}  
}

\clearpage
\newpage
\appendix

\begin{figure*}[h!]
\centering
\begin{subfigure}{0.3\textwidth}
\includegraphics[width=\textwidth]{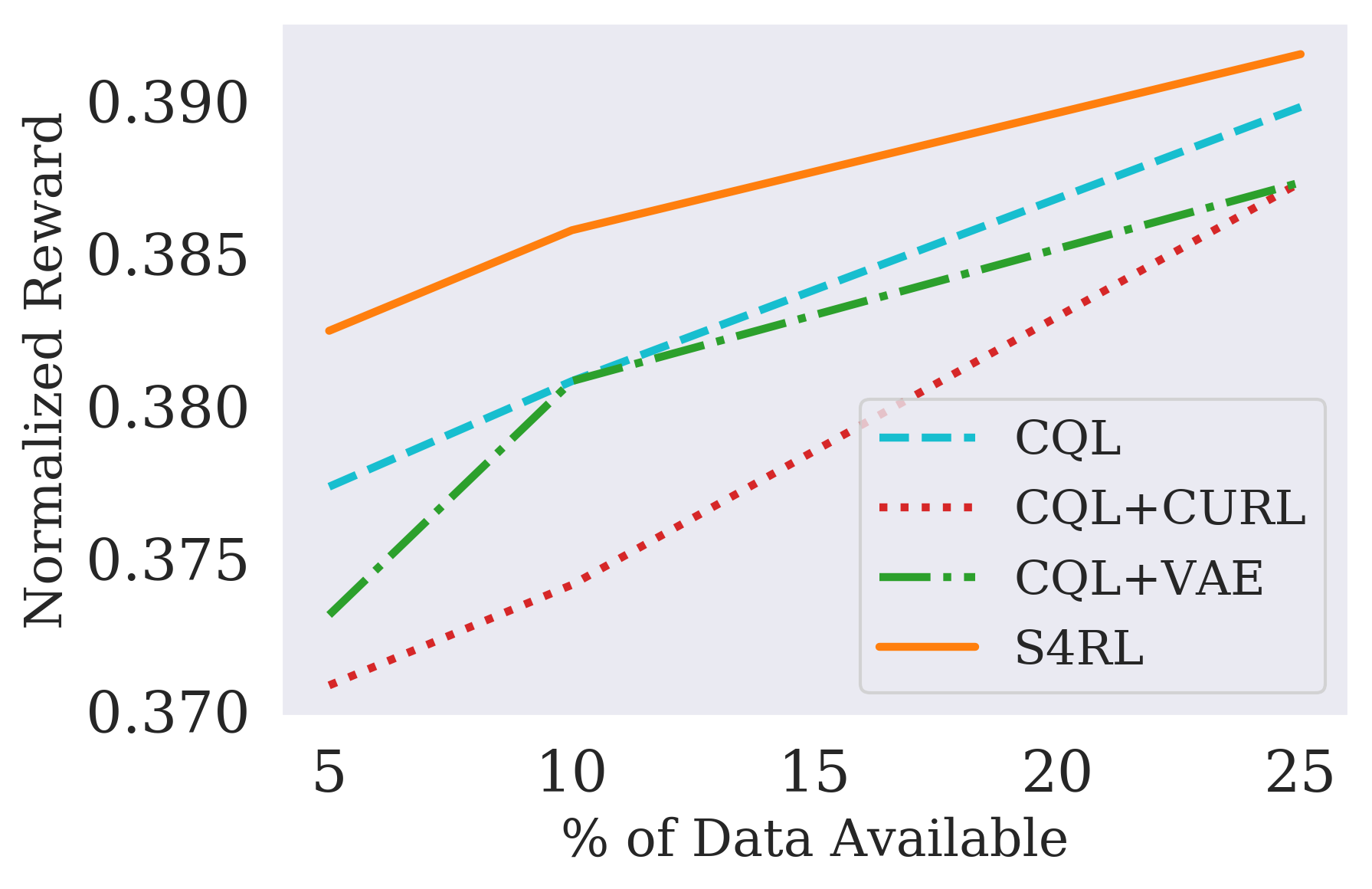}
\caption{cheetah-medium}
~
\end{subfigure}
\begin{subfigure}{0.3\textwidth}
\includegraphics[width=\textwidth]{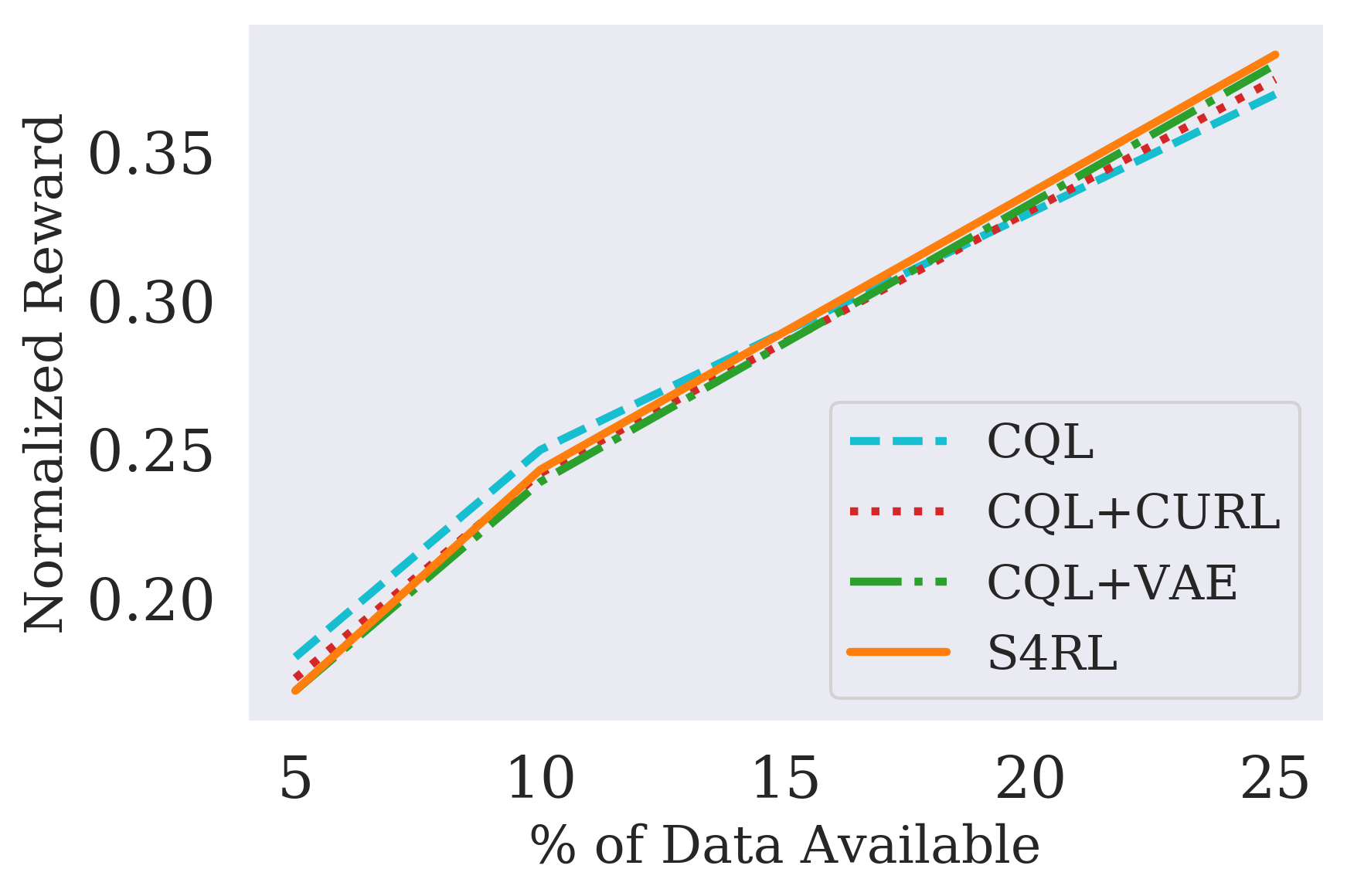}
\caption{cheetah-medium-replay}
\end{subfigure}
~
\begin{subfigure}{0.3\textwidth}
\includegraphics[width=\textwidth]{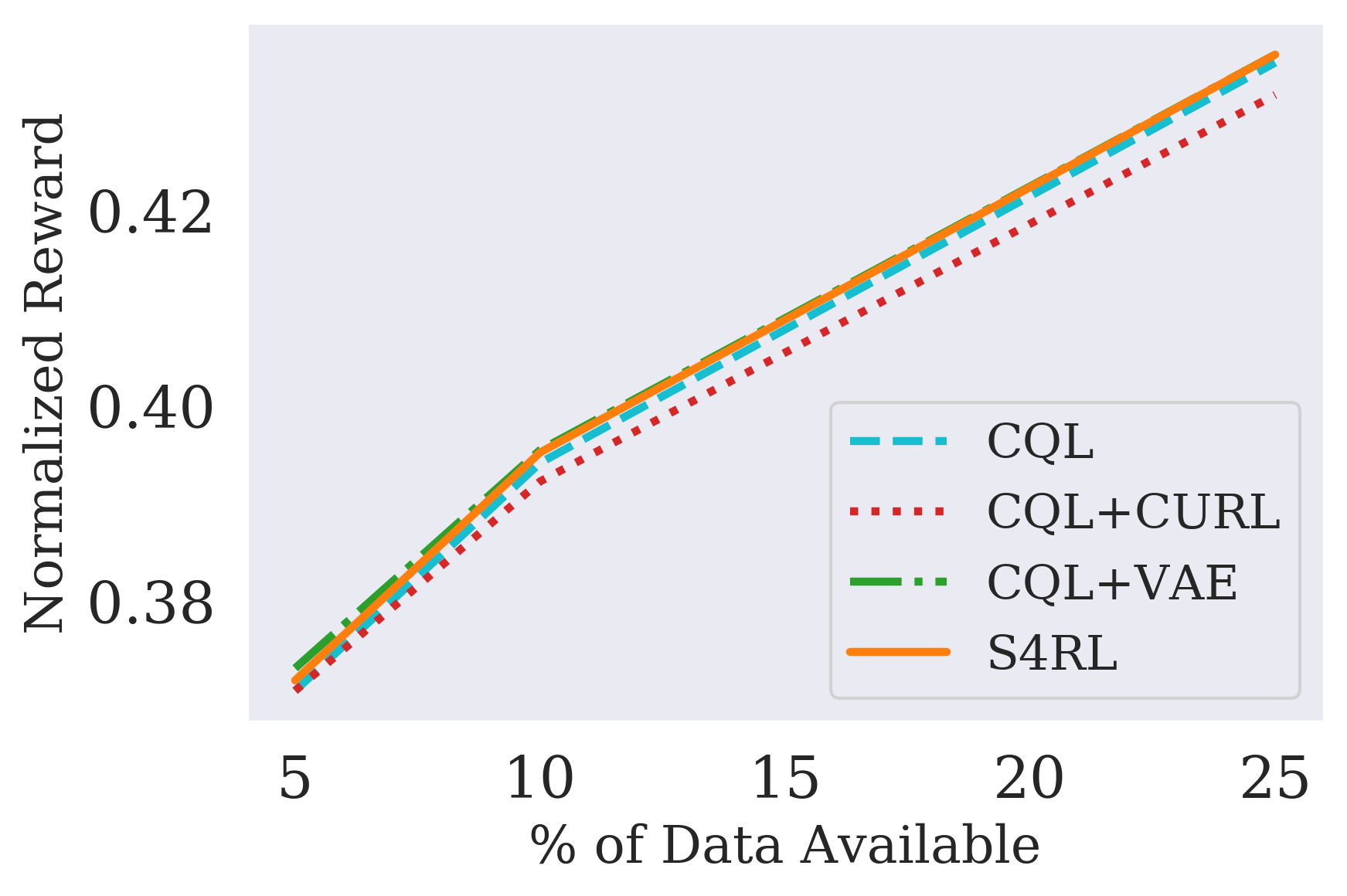}
\caption{cheetah-medium-expert}
\end{subfigure}
~
\begin{subfigure}{0.3\textwidth}
\includegraphics[width=\textwidth]{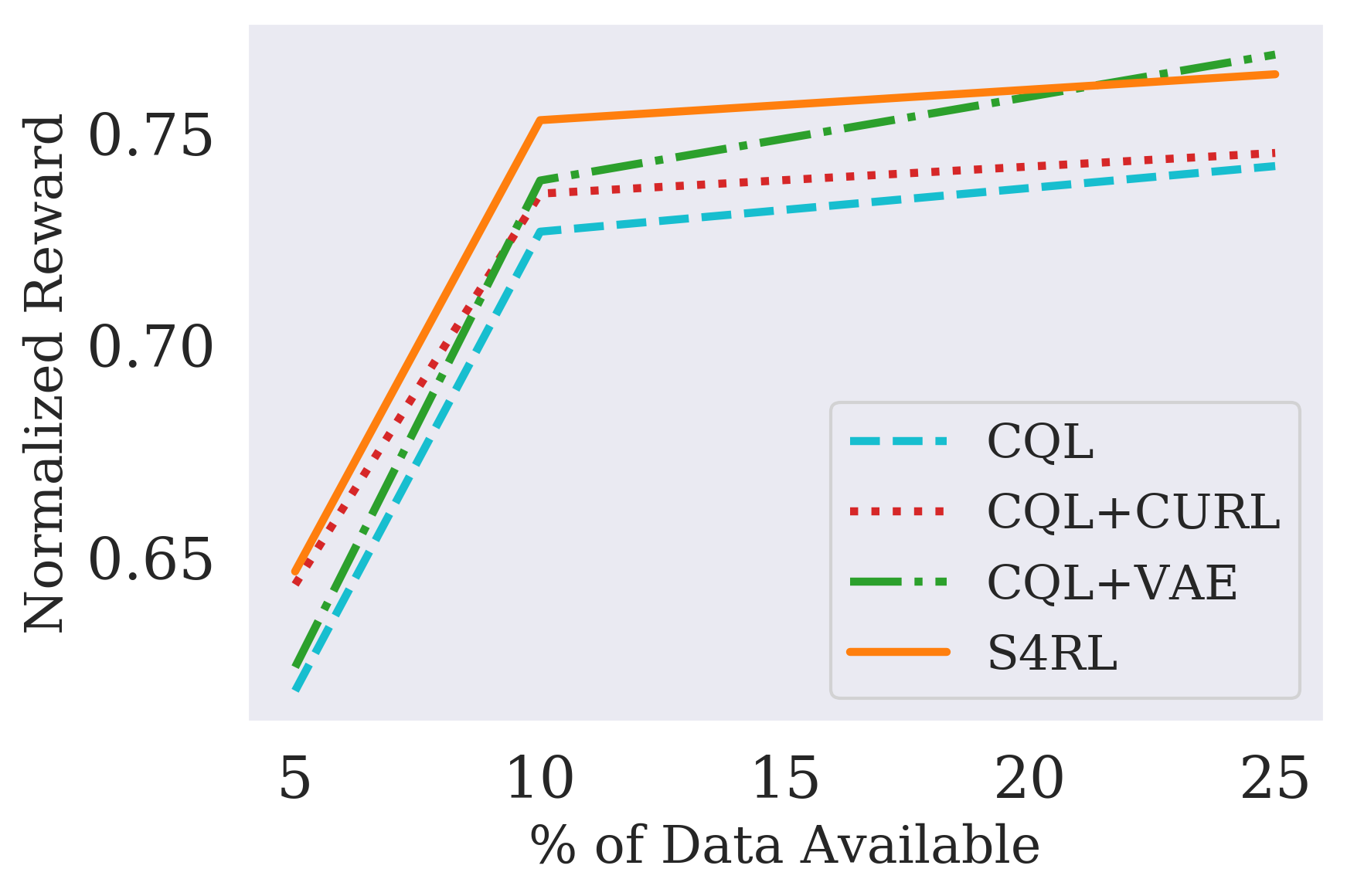}
\caption{hopper-medium}
\end{subfigure}
\begin{subfigure}{0.3\textwidth}
\includegraphics[width=\textwidth]{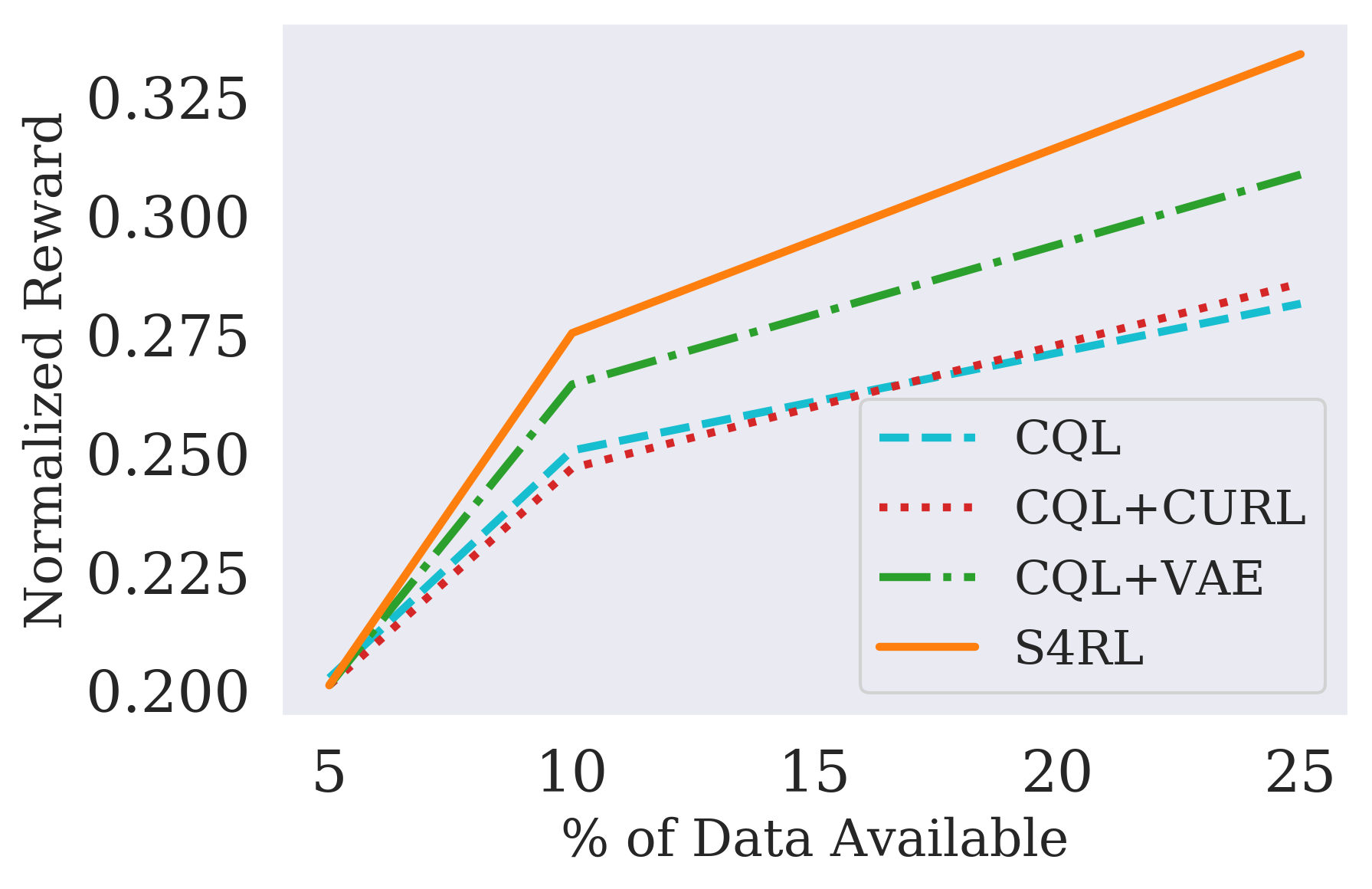}
\caption{hopper-medium-replay}
\end{subfigure}
~
\begin{subfigure}{0.3\textwidth}
\includegraphics[width=\textwidth]{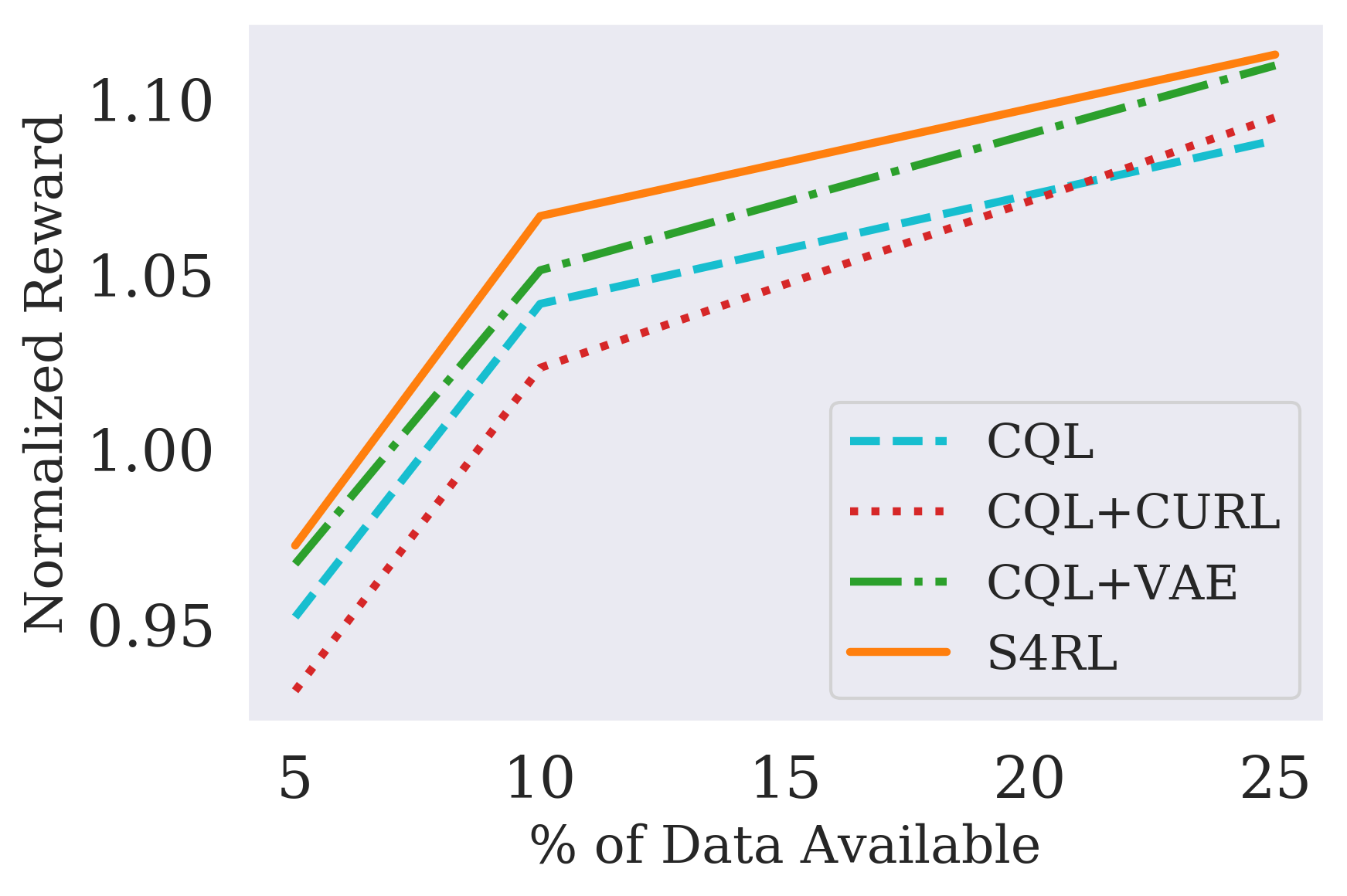}
\caption{hopper-medium-expert}
\end{subfigure}
~
\begin{subfigure}{0.3\textwidth}
\includegraphics[width=\textwidth]{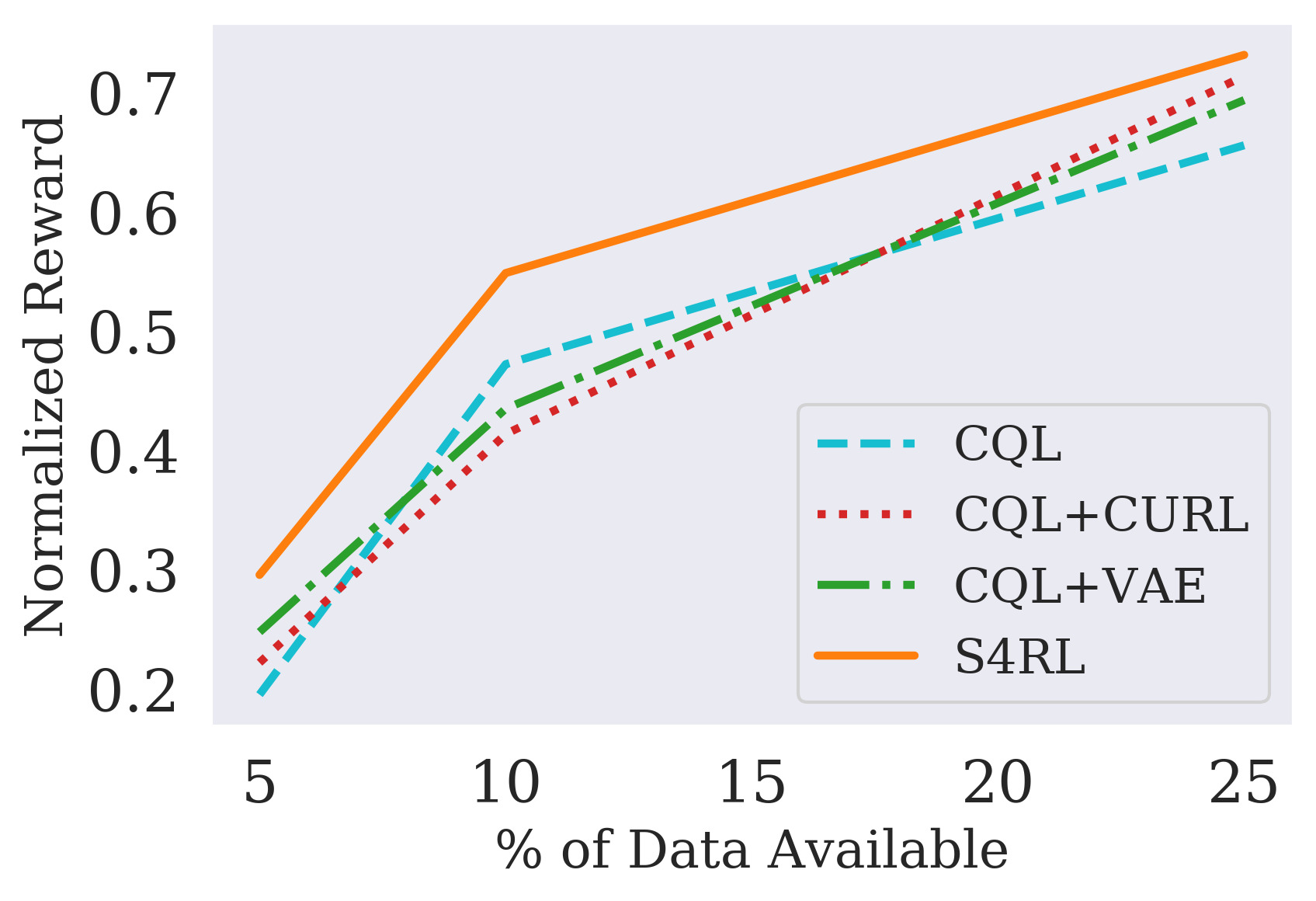}
\caption{walker2d-medium}
\end{subfigure}
\begin{subfigure}{0.3\textwidth}
\includegraphics[width=\textwidth]{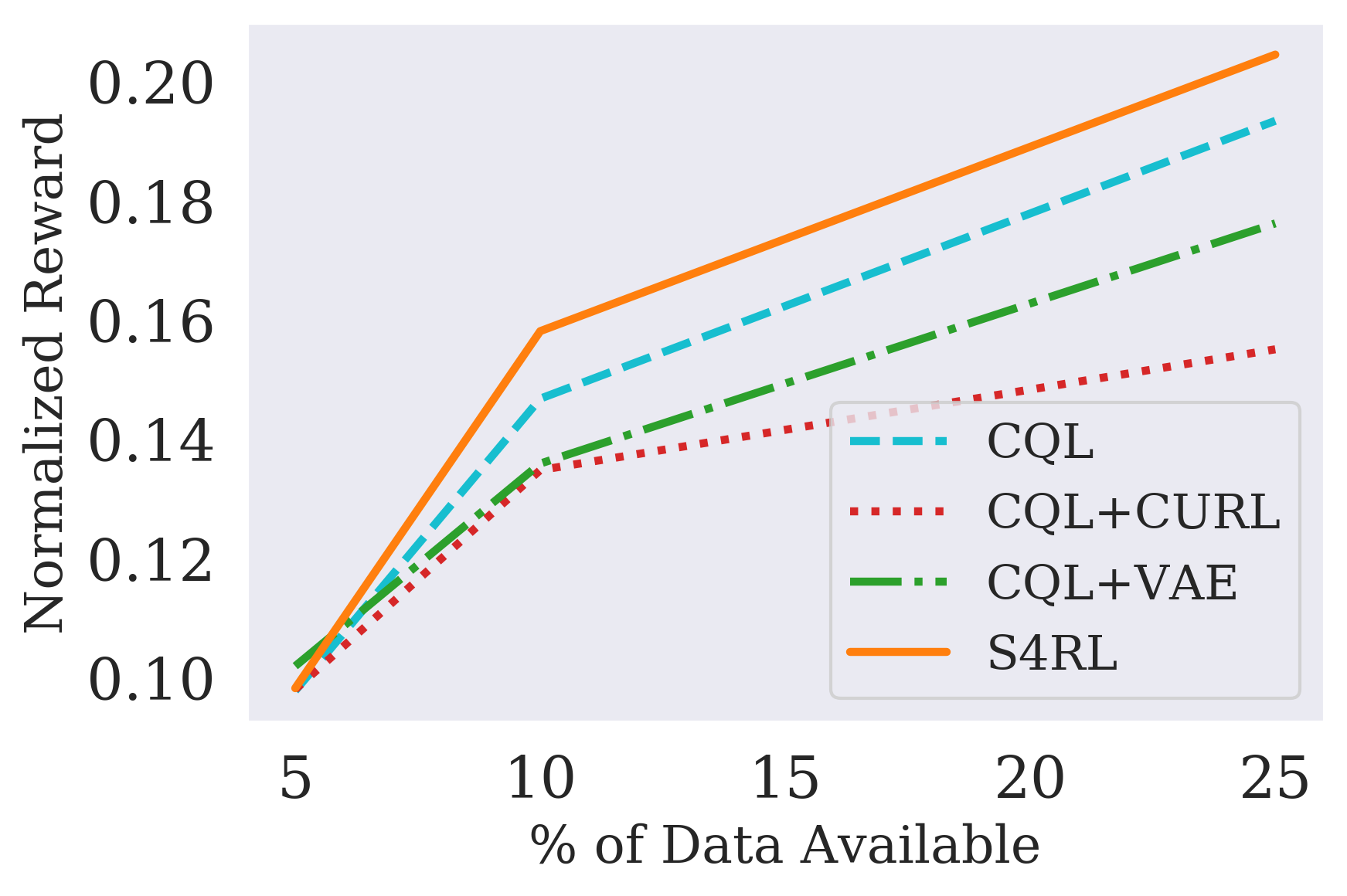}
\caption{walker2d-medium-replay}
\end{subfigure}
~
\begin{subfigure}{0.3\textwidth}
\includegraphics[width=\textwidth]{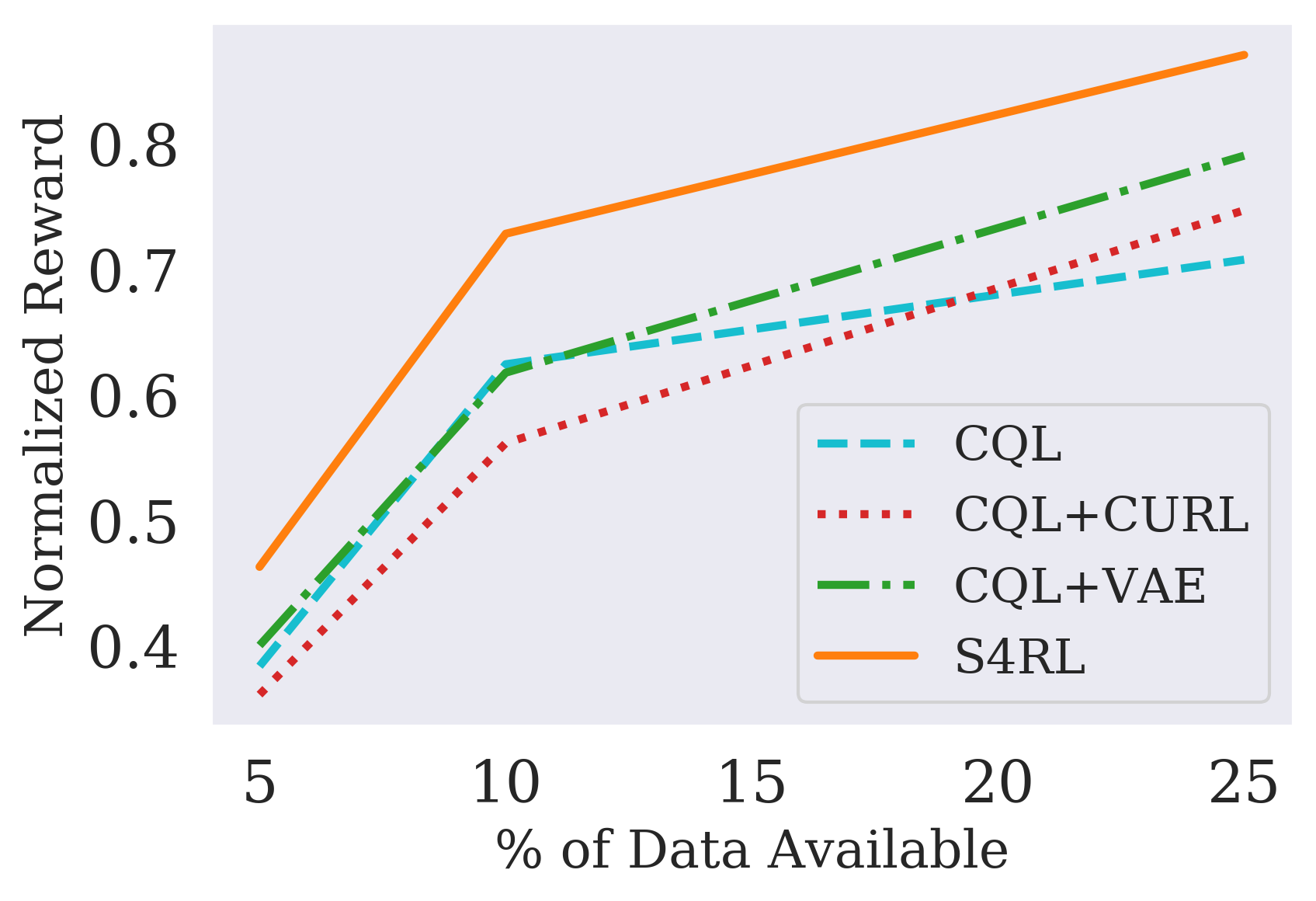}
\caption{walker2d-medium-expert}
\end{subfigure}
\caption{\textbf{Learning from limited data.} 
We train an S4RL agent and the baselines on a small $5\%, 10\%,$ and $25\%$ 
of the different D4RL data splits.
We see that S4RL is consistently the best performing algorithm when learning
from limited data.}
\label{fig:limited_data}
\end{figure*}

\section{Learning with limited data}
\label{sec:limited_data}
Its possible that only a small amount of offline data is available to train
an offline agent, which motivates us to test the agents on learning from 
limited data.
We do this by randomly selecting 5\%, 10\%, and 25\% of the datasets in each
data split for the 3 OpenAI Gym tasks on 3 different data splits:
``-medium'', ``-medium-replay'' and ``-medium-expert''.
By doing so, we significantly limit the amount of information the agent has 
about the environment thereby increasing the difficulty of learning a good
representation from data.
We present the normalized reward (y-axis) over the different data \% available 
(x-axis) in Figure \ref{fig:limited_data}.
On the Hopper and Walker2d data splits, the S4RL-$\mathcal{N}$ agent is able to learn 
significantly outperform all the baselines on both environments in each data split 
and data-percent.
Despite the simplicity and ease of implementation, the agent is able to learn
better representations than CQL+CURL and CQL+VAE.

\newpage

\section{Full Robotic Experiments}
\label{sec:robotic_table}

\begin{table}[!h]
    \centering
    \resizebox{\columnwidth}{!}{%
        \begin{tabular}{l|l|c || c c c c c}
        \toprule
        & & \textbf{Behaviour} & & & & \textbf{CQL} & \textbf{CQL} \\
        \textbf{Domain} & \textbf{Task} & \textbf{policy at 1M} & \textbf{CQL} & \textbf{CQL+CURL} & \textbf{CQL+VAE} & \textbf{+S4RL-$\mathcal{N}$} &\textbf{+S4RL-Adv} \\
        \midrule
        \multirow{5}{*}{MetaWorld} & reach-v1 & 95\% & 77\% & 78\% & 79\% & \textbf{84\%} & \textbf{93\%} \\
        & reach-wall-v1 & 90\% & 54\% & 57\% & 67\% &  \textbf{78\%} & \textbf{88\%} \\
        & sweep-into-v1 & 75\% & 32\% & 33\% &  41\% & \textbf{63\%} & \textbf{\textcolor{Blue}{79\%}} \\
        & door-close-v1 & 35\% & 21\% & 25\% & \textbf{30\%} & \textbf{30\%} & \textbf{\textcolor{Blue}{39\%}} \\
        & push-v1 & 30\% & 12\% & 12\% & 18\% & \textbf{27\%} & \textbf{\textcolor{Blue}{35\%}} \\
        \midrule
        \multirow{2}{*}{RoboSuite} & block-lifting & - & 28\% & 28\% & 37\% & \textbf{45\%} & \textbf{53\%} \\
        & pick-and-place-can & - & 20\% & 16\% & 29\% & 33\% & \textbf{44\%}\\
        \bottomrule
        \end{tabular}%
    }
    \caption{Experiments on robotic manipulation in the Metaworld environment \cite{metaworld}.
    We report the \% goals completed by the behaviour policy at 1M steps for MetaWorld environments.
    When the offline policy outperforms the behaviour policy, we highlight that 
    in \textbf{\textcolor{Blue}{blue text}} and when the S4RL agent outperforms
    the baselines we highlight it in \textbf{bold}.
    These are the numerical results used in Figure \ref{fig:robotic_exps}.
    }
    \label{tab:metaworld_table}
\end{table}

\section{$Q$-network regularization}
\begin{table*}[h!]
    \centering
        \begin{tabular}{l || c c c c c}
        \toprule
         & &  \textbf{CQL}  &  \textbf{CQL}  & \textbf{CQL} \\
        \textbf{Task} & \textbf{CQL} & \textbf{+Dropout} & \textbf{+L2-Reg} &\textbf{+S4RL-Adv} \\
        \midrule
        reach-v1 & 77\% & 68\% & 79\% & \textbf{93\%} \\
        reach-wall-v1 & 54\% & 43\% & 51\% & \textbf{88\%} \\
        sweep-into-v1 & 32\% & 22\% &  34\% & \textbf{{79\%}} \\
        door-close-v1 & 21\% & 18\% & 21\% & \textbf{{39\%}} \\
        push-v1 & 12\% & 4\% & 9\% & \textbf{{35\%}} \\
        \bottomrule
        \end{tabular}%
    \caption{Effect of different regularization schemes on the performance
    over 5 difficult robotic tasks in the MetaWorld environments \cite{metaworld}.
    We use $p=0.3$ for Dropout and weight decay value of $1 \times 10^{-4}$ for 
    L2-Regularization.}
    \label{tab:regularization_table}
\end{table*}

\label{sec:regularization}
Along with self-supervision, in deep learning literature there exist other forms of 
explicit regularization methods that are used to help with performing better 
function approximation by preventing overfitting. 
We include experiments with two such methods:
Monte Carlo dropout mask (MC-Dropout) \cite{dropout} and L2-regularization over  $Q$-network weights. The results are shown in Table 
\ref{tab:regularization_table}.

\section{Ablation over Hyperparamaters}
\label{sec:hyperparam}

\begin{table}[th!]
    \centering
        \begin{tabular}{l| c c c | c c c}
        \toprule
        & \multicolumn{3}{c}{\textbf{CQL+S4RL-$\mathcal{N}$}} & \multicolumn{3}{|c}{\textbf{CQL+S4RL-Adv}} \\
        \textbf{Task} & \textbf{($\sigma=10^{-2}$)} & \textbf{($\sigma=10^{-3}$)} & \textbf{($\epsilon=10^{-4}$)} & \textbf{($\epsilon=10^{-2}$)} & \textbf{($\epsilon=10^{-3}$)}& \textbf{($\epsilon=10^{-4}$)} \\
        \midrule
        reach-v1 & 53\% & 80\% & \textbf{85\%} & 50\% & 88\% & \textbf{93\%} \\
        reach-wall-v1 & 49\% & 70\% & \textbf{72\%} & 44\% & 79\% & \textbf{88\%} \\
        sweep-into-v1 & 47\% & \textbf{64\%} & 61\% & 36\% & 65\% & \textbf{{79\%}} \\
        door-close-v1 & 8\% & 23\% & \textbf{26\%} & 19\% & 32\% & \textbf{{39\%}} \\
        push-v1 & 5\% & 22\% & \textbf{27\%} & 13\% & 27\% & \textbf{{35\%}} \\
        \bottomrule
        \end{tabular}
    \caption{Hyperparameter study on the Metaworld environment \cite{metaworld}.
    We see that S4RL performs best when only local perturbations are applied to the
    states.
    As noted in Section \ref{sec:exp_setup}, we use $\sigma=3\times 10^{-4}$ and
    $\epsilon=1 \times 10^{-4}$ for S4RL-$\mathcal{N}$ and S4RL-Adv, respectively.
    }
    \label{tab:hyperparam_table}
\end{table}

The results for ablation over values of $\sigma$ and $\epsilon$ for the two
best performing methods (S4RL-$\mathcal{N}$ and S4RL-Adv) are presented in
Table \ref{tab:hyperparam_table}.
We see that large values of $\epsilon$ or $\sigma$ ($10^{-2}$)significantly 
hurt the 
performance of S4RL, whereas the model is relatively stable within the 
range of $10^{-3}$ and $10^{-4}$ suggesting the need for local perturbations
to the proprioceptive information. 
A large variance or a large adversarial sample hurts the performance since the
new states $T(s_t)$ may be semantically different than the original state,
since proprioceptive inputs are continuous unlike pixel-based input.

\section{Implementation details on state-switch}
\label{state-switch}

The implementation of the state-switch experiments is done by using ad-hoc rules
for each environment where we only replace dimensions which are similar to each
other.
Such as a joint-angle is only replaced by another angle, and a joint velocity is 
only replaced by another velocity. 
Using this scheme ensures that we remain approximately in the same
bounds of what is physically realizable, since its possible that velocities and
angles do not work on the same scale.
Since state-switch needs significantly more oracle knowledge of the environment
and the state space, it may not be the best choice of augmentation for environments
where the state space is not known.
Furthermore, we see that in practice, state-switch is unable to perform well on most 
baselines as shown in Table \ref{tab:aug_table}, which consolidates the relative
ineffectiveness of the augmentation choice.

\section{Details about Robosuite experiments}
\label{sec:robosuite_data}

Robosuite~\cite{robosuite} datasets were collected by first training a Soft Actor Critic~\cite{haarnoja2018sac} RL agent from scratch on the Lift and Can tasks. Agent checkpoints were saved regularly during training -- 5 checkpoints for the Lift task, and 17 checkpoints for the Can task. For each checkpoint, 300 agent rollouts were collected (with horizon 150) for a total of 1500 Lift trajectories and 5100 Can trajectories. Consequently, these datasets contain a mixture of expert and suboptimal trajectories, and resemble datasets from common offline RL benchmarks~\cite{d4rl, d4pg}.

\end{document}